# High Voltage Insulator Surface Evaluation Using Image Processing

Damira Pernebayeva, Mehdi Bagheri, and Alex James
Electrical and Computer Engineering Department
Nazarbayev University
damira.pernebayeva@nu.edu.kz

*Abstract* –High voltage insulators are widely deployed in power systems to isolate the live- and dead-part of overhead lines as well as to support the power line conductors mechanically. Permanent, secure and safe operation of power transmission lines require that the high voltage insulators are inspected and monitor, regularly. Severe environment conditions will influence insulator surface and change creepage distance. Consequently, power utilities and transmission companies face significant problem in operation due to insulator damage or contamination. In this study, a new technique is developed for real-time inspection of insulator and estimating the snow, ice and water over the insulator surface which can be a potential risk of operation breakdown. To examine the proposed system, practical experiment is conducted using ceramic insulator for capturing the images with snow, ice and wet surface conditions. Gabor and Standard deviation filters are utilized for image feature extraction. The best achieved recognition accuracy rate was 87% using statistical approach the Standard deviation.

*Keywords: insulator, image processing, inspection technique, Gabor filter, Standard deviation, k-NN classifier*

## I. INTRODUCTION

In severe cold climate conditions, the properties of electrical high-voltage overhead insulator surface tend to deteriorate due to the accumulation of ice and snow in combination with strong wind. There are two main concerns related to icing effect: the weight of ice and the conductivity of water. The accumulated ice leads to flashover as it reduces dramatically the insulator leakage distance [1]. Icing related insulator faults have proved to result in power outages across the world including Canada, Norway, Japan, the US and UK [1]. The timely inspection of insulator surface is an important procedure to identify the early stage of aging and deformation occurring in severe environmental conditions.

This study uses the Gabor and Standard deviation filters with the k-NN classifier to assess the condition of insulator surface by determining the present of ice, snow and water using images. These two filters are applied to extract the features from the images of insulator surface. The images are prepared using the ceramic insulator under winter conditions: snow, ice and melt water. The database was created by acquiring the images with three classes. The k-NN classifier is utilized to assess the classification accuracy of identifying nature of insulator surfaces. Results obtained from sumalitions show the potential of use of image filtering techniques to extract useful features for assess the insulator surfaces.

This paper is organized as follows. In section II insulator types and insulator inspection methods are discussed. In section III the proposed system is introduced including: Image analysis and Classification, Feature extraction, and the k-nearest neighbour classifier. Results and Analysis are discussed in section IV and the conslusion is given in section V.

## II. TYPES OF INSULATOR AND INSPECTION TECNIQUES

### A. Types of Insulators

High Voltage (HV) insulators provide mechanical support for the transmission line conductors and prevent the current flow from conductors to the earth. HV insulators are classified based on how they supposed to be used on power lines. Their classification can be based on their electrical, mechanical and environmental service stresses.

Ceramic, toughened glass, and polymers are commonly used materials for high voltage insulator fabrication. Desirable resistivity is considered as the main advantage in ceramic and glass insulators, however, poor hydrophobicity vulnerability to fend contamination are considered as the main disadvantages. In contrast, the polymer insulators are lighter than ceramic insulators with higher surface hydrophobicity. However, they tend to age fast and have lower withstanding to mechanical loads compared to ceramic insulators.

There are several types of insulators such as pin type insulator, post insulators, suspension type and strain insulators that are widely deployed in electric power distribution system [2]. Characteristics of listed insulators are briefly presented in Table I.

TABLE I. INSULATOR TYPES [2]

| Insulator type | Operating Voltage | Device description |
|---|---|---|
| Pin type insulators | less than 33kV | Economic and efficient; it can withstand against wind, and mechanical stress |
| Post insulator | up to 33 kV | Used for supporting bus bars and disconnecting switches in sub stations |
| Suspension insulator | each unit designed for 11 kV | Consists number of porcelain discs connected in series in form of a string; the number of insulator discs in a string depends on operating voltage. It is cheaper for operating voltage above 33kV, less mechanical stress. It is cheaper for operating voltage above 33kV, less mechanical stress. |

## B. Insulator monitoring methods

The regular checking of insulators on overhead lines is quite a challenge due to the height of towers, location of power lines (mountains, terrains and other non-accessible places) and voltage applied across the insulators. Therefore, researchers have recently focused on the development of remote and automated inspection techniques using signal processing. The various methods, based on laboratory experiments and in-the-field tests, have been proposed to inspect the insulator. In [3], the Hough transform line detection method has been proposed to study the insulator tilt affects using morphological operations. The fault detection accuracy was performed by using feature extraction technique the Gray-level-Co-occurrence matrices (GLCM) form airborne images. There is also an attempt to use fast Fourier transform (FFT)-based spectral analysis to examine the polluted surface condition of insulator [4]. Another work evaluates the hydrophobicity of the insulator surface using the co-occurrence matrix to extract the features from the binary images [5]. The remote condition monitoring of insulator has been proposed by [6] where the adaptive neuro-fuzzy inference system and SMV have been used to determine the condition of the insulator.

Table II describes the summary review on the insulator diagnostic methods with their advantages and limitations.

TABLE II. INSULATOR INSPECTION METHODS

| Insulator inspection Methods | Advantages | Limitations |
|---|---|---|
| A computational electric field approach [7] | evaluates the condition of an insulator and detect internal defects | not accurate for internal cracks |
| Infrared and ultraviolet imaging technique [8] | detects the temperature variance caused by internal cracks | high cost, limited number of conditional defects; sensitive to the environment |
| Harmonic retrieval method [3] | remote inspection based on analysis of the low frequency even-order harmonic of the basic power line transmission frequency (60 Hz), can be used in the high voltage power lines | high cot, time consuming |
| Watershed algorithm technique [9] | detection of the dirt area in the images of the insulator surface | sensitive to small changes, over-segmentation phenomena |
| Hough transform line detection method [3] | insulator tilt correction based on Hough transform technique; stable and robust to noises in images | error in tilt angle calculation |
| Insulator Hydrophobicity inspection [5] | the polynomial classifier performs high classification accuracy | applicable for a small class number due to its computation complexity. |

## III. IMAGE ANALYSIS AND CLASSIFICATION

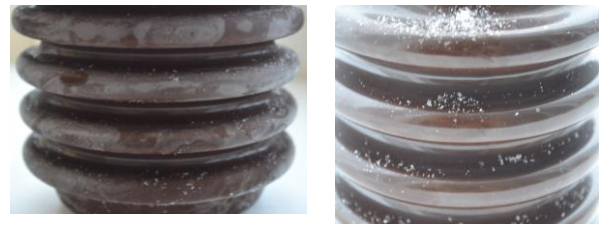

With ice    With snow

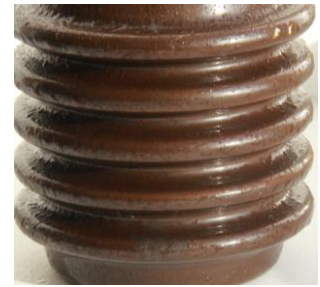

with sprayed water

Figure 1. Insulator image samples

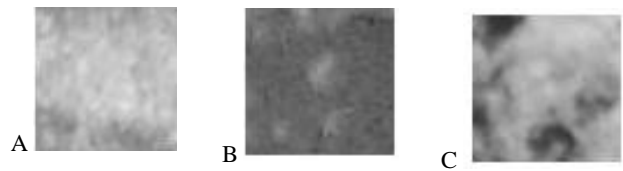

A    B    C

Figure 2. Segmented: A. snow, B. water, C ice

The insulator inspection through the analysis of images is performed using the set of steps shown in in Fig. 2. In the image acquisition stage, a database of insulator images covered by snow, ice and sprayed water are prepared. This dataset contains of 377 images with 3 classes. To reduce the storage computational complexity, the original RGB images obtained by the camera were downsized by cropping square segments of size 64x64 pixels and converted into grayscale. The data set was divided into train and test set, selected randomly, for testing the classification accuracy. The samples of the insulator covered with snow, ice and water are illustrated in Fig. 1.

### A. Feature extraction

Feature extraction is the process of identifying useful features that can often reduce the dimension of features using a feature transform. The extracted features are expected to contain discriminatory information of the input data useful for classification purposes. For this work, a comparison is drawn between two different biologically inspired techniques like Gabor and Standard deviation filters. Two techniques for image analysis and feture extraction were used.

Gabor filter has been known as useful tool in image processing due to its similarity with the human visual system. In general, Gabor filter is a linear filter used for image texture and edge detection analysis. Basically, it is a two dimensional, frequency and orientation sensitive band pass filter, formed by sinusoids and Gaussians function *(1)*. It has been widely used in many different applications such as: edge and object detection, fingerprint recognition, biometrics, medical and biological applications [10, 11, 12]. In general, the principal work of the filter is that image patterns are convolved with the Gabor filters using different preferred spatial frequencies and orientations to evaluate their similarities. Each wavelet captures energy at a specific frequency and direction which provide a localized frequency as a feature vector

$$F(x, y) = -\exp\left(\left\{\frac{1}{2}\left[\frac{x_\theta^2}{\sigma_\theta^2} + \frac{y_\theta^2}{\sigma_\theta^2}\right]\right\}\cdot \cos(2\pi \cdot f \cdot x_\theta)\right) \quad (1)$$

$$x_\theta = x\cdot\cos\theta + y\cdot\sin\theta$$
$$y_\theta = -x\cdot\sin\theta + y\cdot\cos\theta$$

where the filter is described by the following parameters $\sigma_x$ and $\sigma_y$ –sigma, represent the standard deviation of Gaussian envelope in *X* and *Y* direction and controls the width of Gaussian envelope, $\theta$ – theta, this is the rotation for the filter related to x-axis, $f$ -is the frequency of a sinusoidal plane wave.

The standard deviation filter is a statistical approach where the filter substitutes the original pixel-value with a standard deviation window mask in an area around the pixel [13]. This operation resembles the human eye sensory filtering process resulting in edges of different strengths determined by the window size. Three window masks 3x3, 5x5, 7x7 are applied to get feature vectors for all images.

$$\sigma = \sqrt{\frac{1}{N}\sum_i^N (x_i - \mu)} \quad (2)$$

### A. Classification

The purpose of classification process is to categorize certain pixels of the feature vectors in digital image into one of class labels. The k-NN classifier has been widely used for statistical estimation and pattern recognition. It basically performs the similarity measurements for the matching the feature vectors of train set images with tested set. The measurement is done by summing up the difference $\Delta S$ between extracted feature $F_q$ of test set and $F_p$ of train set *(3)* [15].

In the experiment, the dataset consists of 377 images divided into 3 class labels including snow, ice and water (Fig. 3). The interest regions in the input images are cropped by the size 64x64 pixels to reduce computation time and divided into train and test set, selected randomly. When the feature vectors are created for all images in database, the k-nearest neighbor (k-NN) distance based classifier algorithm is used.

$$\Delta S = \sum_i^n \left|(F_p(i) - F_q(i))\right|, \text{ where i=1, 2, .., n} \quad (3)$$

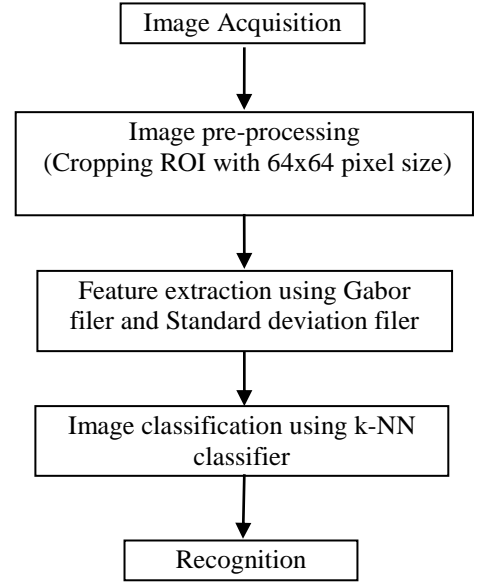

Figure 3. Flowchart of image recognition steps performed

## IV. RESULTS AND ANALYSIS

In this study, two feature extraction algorithms are successfully tested using the insulator surface image dataset, which consists of 377 images with 3 classes: ice, snow and water (Fig. 3). In Fig. 4 and 5, the error bar graphs represent the performance of the k-NN classifier based on the extracted features using the Gabor and standard deviation filters, respectively. The graphs are constructed using mean and standard deviation. In Fig.4 and 5, the standard deviation increases with the increase of the number of training samples. Basically, standard deviation takes square root of variance, which measures the distance between points. The recognition accuracy using the standard deviation filter performs better results than the Gabor filter, 86% and 59%, respectively.

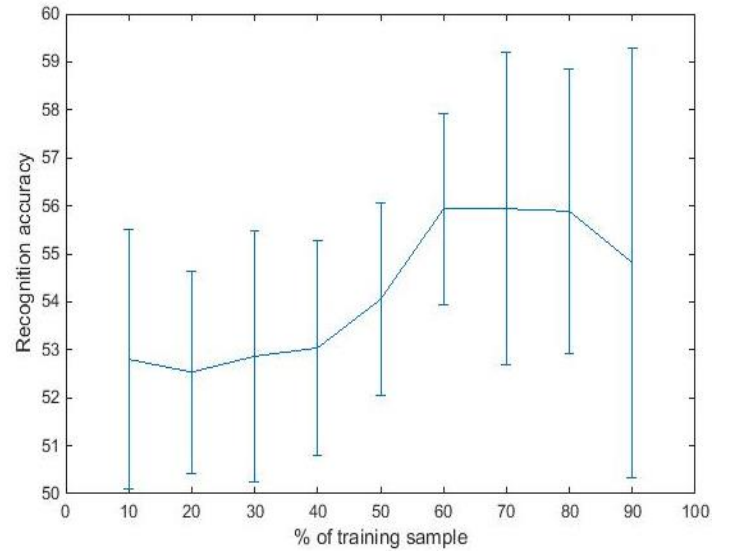

Figure 4. Recognition accuracy for three classes using Gabor filter

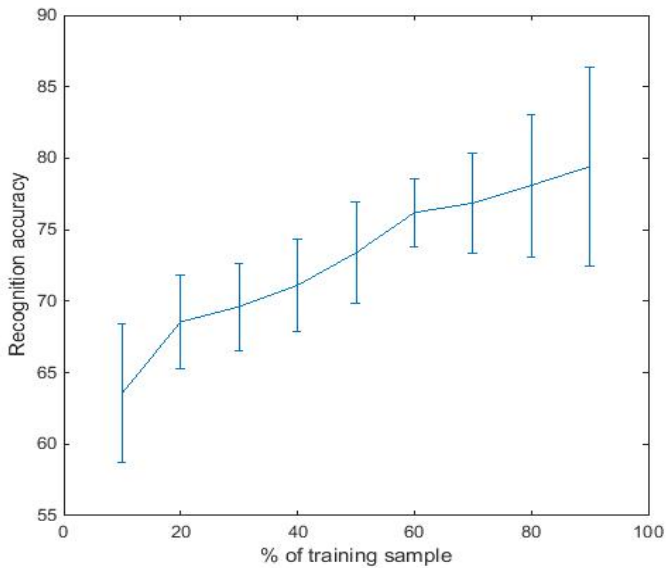

Figure 5. Recognition accuracy for three classes using Standard deviation

## V. CONCLUSION

In this paper two feature extraction techniques including the Gabor and Standard deviation filters were examined for insulator surface evaluation. The obtained feature vectors were utilized using the k-nearest neighbor classifier. The results of the system were based on 377 images acquired and pre-processed from the surface of the ceramic insulator under winter conditions. Three classes with surface covered with snow, ice and water were used to estimate the recognition accuracy of the system. The recognition accuracy based on standard deviation performs better results comparing to Gabor filter.

REFERENCES

[1] "Coating for protecting overhead power power network equipment in winter conditions", 2015.
[2] Zhang, X., and Chen, F. " A Method of Insulator Fault Detection from Airborne Images", Second WRI Global Congress on Intelligent Systems, 2010.
[3] Han, S., Hao, R., Lee, J., & Member, S. "Inspection of Insulators on High-Voltage Power Transmission Lines", IEEE Transactions on Power Delivery, 2009, pp 2319–2327.
[4] Chandrasekar, C. M. S. "Adaptive neuro-fuzzy inference system for monitoring the surface condition of polymeric insulators using harmonic content" IET Generation, Transmission and Distribution, 2010, pp751–759.
[5] Wang, Z., Yu, Z., Xu, J., Wang, X., Yu, H., Zhao, D., & Han, D.). "High Voltage Transmission Lines Remote Distance Inspection System", IEEE International Conference on Mechatronics and Automation 2015,pp 1676–1680.
[6] Kontargyri, V. T., Gonos, I. F., Stathopulos, I. A., & Alex, M. "Simulation of the Electric Field on High Voltage Insulators using the Finite Element Method", 12th Biennial IEEE Conference on Electrical Field Computation, 2006,.
[7] Fang-cheng, L., Hu, J., Sheng-hui, W., & Hai-de, L. (2012). "Fault Diagnosis Based on Ultraviolet Imaging Method on Insulators", 1426–1429.
[8] Khalayli, L., Sagban, H. Al, Shoman, H., Assaleh, K., & El-hag, A. (2013). Automatic Inspection of Outdoor Insulators using Image Processing and Intelligent Techniques, Electrical Insulation Conference, 2013,pp 206–209.
[9] Xin, M., Tiecheng, L., Xiaoyun, W., & Bo, Z. "Insulator Surface Dirt Image Detection Technology Based on Improved Watershed Algorithm",Asia-Pacific Power and Energy Engineering Conference, 2012.
[10] Kamarainen, J.-K., Kyrki, V., & Kalviainen, H. (2006). Invariance properties of Gabor filter-based features-overview and applications. IEEE Transactions on Image Processing, 15(5), 1088–1099.
[11] Liu, S. L., Niu, Z. D., Sun, G., & Chen, Z. P. (2014). Gabor filter-based edge detection: A note. *Optik*, *125*(15), 4120–4123.
[12] Liu, C. L., Koga, M., & Fujisawa, H. (2005). "Gabor feature extraction for character recognition: Comparison with gradient feature." Proceedings of the International Conference on Document Analysis and Recognition, ICDAR, 2005, 121–125.
[13] Technology, A. I., & Baharudin, B. Bin. "Mean and standard deviation featurs of color historgram using laplacian filter for content based image retrieval", Journal of Theoretical and Applied Information Technology *34*(1), 2011.
[14] Randen,T, J.H Husoy. "Filtering for Texture Classification : A Comparative Study", IEEE Transactions on Pattern Analysis and Machine Intelligence, 1999, pp291–310.